\documentclass[conference]{IEEEtran}
\IEEEoverridecommandlockouts
\usepackage{cite}
\usepackage{amsmath,amssymb,amsfonts}
\usepackage{algorithmic}
\usepackage{graphicx}
\usepackage{textcomp}
\usepackage{xcolor}
\usepackage{amsmath,epsfig}
\usepackage{indentfirst}
\usepackage{booktabs}
\usepackage{colortbl} 
\usepackage{float}
\usepackage{multirow}
\usepackage{cite}
\usepackage{amsmath,amssymb,amsfonts}
\usepackage{algorithmic}
\usepackage{graphicx}
\usepackage{textcomp}
\usepackage{marvosym}

\def\BibTeX{{\rm B\kern-.05em{\sc i\kern-.025em b}\kern-.08em
    T\kern-.1667em\lower.7ex\hbox{E}\kern-.125emX}}
\begin{document}

\title{Exploring State Space Model in Wavelet Domain: An Infrared and Visible Image Fusion Network via Wavelet Transform and State Space Model
\thanks{\textsuperscript{\textdagger} Equal Contribution. \textsuperscript{\Letter} Corresponding author. This research was supported by the Zhejiang Provincial Natural Science Foundation of China (Grant No. LY22F050002) and the Graduate Scientific Research Foundation of Hangzhou Dianzi University (Grant No. CXJJ2024070).}
}

\author{\IEEEauthorblockN{Tianpei Zhang\textsuperscript{1,\textdagger},Yiming Zhu\textsuperscript{1,\textdagger},Jufeng Zhao\textsuperscript{1,\Letter},Guangmang Cui\textsuperscript{1},Yuchen Zheng\textsuperscript{1}}
\IEEEauthorblockA{\textit{\textsuperscript{1}School of Electronics and Information Engineering, Hangzhou Dianzi University,} Hangzhou, China \\
232040138@hdu.edu.cn, yiming\_zhu\_hdu@163.com, dabaozjf@hdu.edu.cn,\\ cuigm@hdu.edu.cn, 232040254@hdu.edu.cn}
}

\maketitle

\begin{abstract}
Deep learning techniques have revolutionized the infrared and visible image fusion (IVIF), showing remarkable efficacy on complex scenarios. However, current methods do not fully combine frequency domain features with global semantic information, which will result in suboptimal extraction of global features across modalities and insufficient preservation of local texture details. To address these issues, we propose Wavelet-Mamba (W-Mamba), which integrates wavelet transform with the state-space model (SSM). Specifically, we introduce Wavelet-SSM module, which incorporates wavelet-based frequency domain feature extraction and global information extraction through SSM, thereby effectively capturing both global and local features. Additionally, we propose a cross-modal feature attention modulation, which facilitates efficient interaction and fusion between different modalities. The experimental results indicate that our method achieves both visually compelling results and superior performance compared to current state-of-the-art methods. Our code is available at https://github.com/Lmmh058/W-Mamba.
\end{abstract}


\section{Introduction}
\label{sec:intro}

Infrared and visible light images exhibit strong complementary characteristics. Specifically, visible images have rich texture details but are sensitive to lighting variations, while infrared images provide thermal information but lack texture details. Therefore, IVIF can effectively integrate the information of visible and infrared images with rich textures and prominent targets, which is not only enhances human visual observation, but also significantly improves advanced computer vision tasks, such as object detection \cite{hu2024smpisd, zhu2024towards}, semantic segmentation \cite{hu2024gradient}, and tracking \cite{yilmaz2006object}.

Currently, IVIF methods mainly rely on convolutional neural networks, including U2Fusion \cite{xu2020u2fusion} and DenseFuse \cite{li2018densefuse}. These methods combine both image-level and feature-level features, but overlook long-range dependencies during feature extraction, which can decrease fusion performance. GAN-based methods, such as GANMcC\cite{ma2020ganmcc}, transform image fusion as an adversarial learning problem, emphasizing global information, but neglecting the interaction of different domains feature. The vision Transformer based method, such as DATFuse \cite{tang2023datfuse}, utilize self-attention to capture long-range dependencies. However, computation of self-attention mechanisms faces the challenge of quadratic complexity. State-space models (SSMs) have emerged as competitive methods, offering linear scalability to capture long-range dependency. Recently, the Mamba framework\cite{gu2023mamba}, such as FusionMamba \cite{xie2024fusionmamba}, have exhibited outstanding performance in image fusion by capitalizing on the linear complexity of SSM and its exceptional visual modeling capabilities.

\begin{figure*}[htbp]
\centering
\includegraphics[width=1.0 \textwidth]{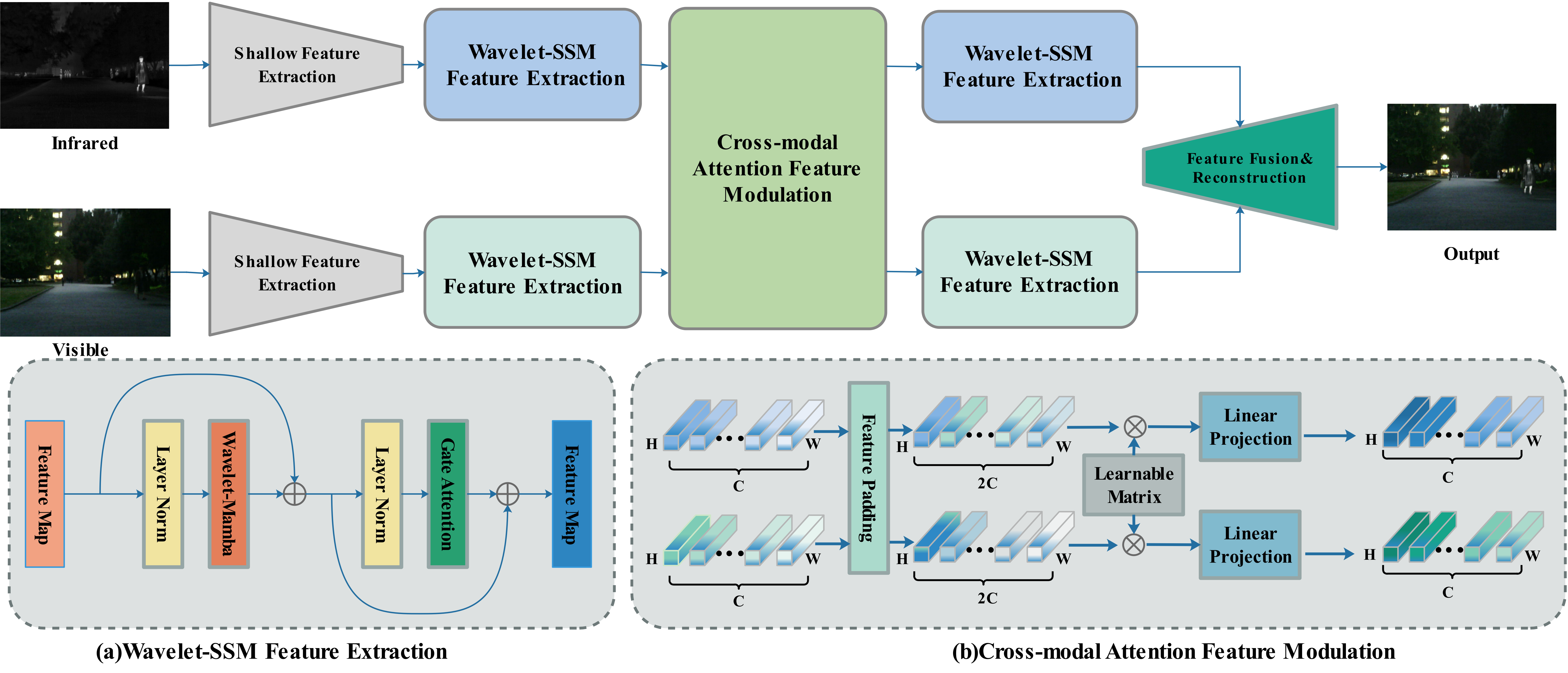}
\caption{The overall structure of the proposed W-Mamba, highlighting the core components of the W-Mamba module. This architecture features two uniquely designed modules: the Wavelet-SSM Feature Extraction (WFE) (a) and the Cross-modal Attention Feature Modulation (CAFM) (b).}
\label{Architecture}
\end{figure*}

In addition to effective long-range dependency modeling, we argue that the key bottleneck of IVIF lies in comprehensive feature extraction across spatial and frequency domains. Recently, the frequency domain fusion method \cite{xiao2024fafusion} achieved promising results by extracting frequency information. However, CNN-based methods \cite{si2022inception} typically focus on high-frequency features while neglecting low-frequency information, while SSM and Transformer excel in capturing long-range dependencies but face challenges in preserving high-frequency features such as edges and textures. Therefore, how to separate high- and low-frequency information and the interaction of frequency-spatial features is remains a problem worth considering. Therefore, IVIF must not only prioritize the global receptive field for feature extraction but also fully integrate the frequency domain features and the information interaction between modalities.

To address the aforementioned problem, we propose Wavelet-Mamba (W-Mamba), which first utilize wavelet transform decomposes image features into low-frequency components that capture structural information and high-frequency sub-bands that contain texture details. Additionally, We combine the advantages of SSM in modeling long-range dependence with the ability of wavelet transform in frequency feature extraction. A cross-domain attention feature modulation was designed to explore intra-inter modal features. This integration enhances feature perception and cross-modal feature interaction in IVIF, improving overall performance. The contributions of this work are as follows:

\begin{itemize}
  \item We propose a novel IVIF framework that comprehensively extracts frequency-domain features, effectively complementing spatial-domain features.
  \item We designed the Wavelet-SSM Feature Extraction module, which extracts both global features and local details from the frequency domain.
  \item We designed a cross-modal feature modulation module that facilitates the efficient interaction of complementary features by fully learning their interdependencies.
\end{itemize}

\section{Method}
\subsection{Overall Framework}
The architecture of Wavelet-Mamba (W-Mamba) is depicted in Fig. \ref{Architecture}. Initially, the source images \(I^{vi}\) and \(I^{ir}\), with dimensions \(R^{C \times H \times W}\), are inputted separately into a CNN-based shallow feature extraction module to extract more informative feature representations.

\begin{equation}
\label{eq:shallow feature}
f_{sf}^{vi} = SF(I^{vi}),f_{sf}^{ir} = SF(I^{ir})
\end{equation}

\noindent where \(f_{sf}^{vi}\) and \(f_{sf}^{ir} \in R^{C' \times H' \times W'}\) \((H' = \frac{H}{2}, W' = \frac{W}{2})\) represent the shallow features extracted by the shallow feature extraction module \(SF(\cdot)\). The source images are downsampled to reduce computational complexity, and their channels are mapped to \(C'\) to enhance the capacity for feature representation. Subsequently, the Wavelet-SSM Feature Extraction Network (WFE) efficiently captures comprehensive frequency-domain information.
\begin{equation}
\label{eq:FIM1}
f_{df}^{vi} = WFE(f_{sf}^{vi}),f_{df}^{ir} = WFE(f_{sf}^{ir})
\end{equation}

\noindent where \(f_{df}^{vi}\) and \(f_{df}^{ir}\) represent the deep features extracted by \(WFE(\cdot)\). The network then performs Cross-Modal Attention Feature Modulation (CAFM) to enhance feature interactions and utilize complementary information, thereby reinforcing the original modality features through an additional WFE.
\begin{equation}
\label{eq:token exchange}
\begin{aligned}
f_{ccf}^{vi} = WFE(CAFM(f_{df}^{vi}))\\
f_{ccf}^{ir} = WFE(CAFM(f_{df}^{ir}))  
\end{aligned}
\end{equation}

\noindent where \(f_{ccf}^{vi}\) and \(f_{ccf}^{ir}\) represent the comprehensive cross-modal features. Finally, \(f_{ccf}^{vi}\) and \(f_{ccf}^{ir}\) are passed through a CNN-based fusion and reconstruction module with upsampling to integrate the features and map them back to the original image space.
\begin{equation}
\label{eq:reconstruction}
I_{F} = F\&R(f_{ccf}^{vi},f_{ccf}^{ir})
\end{equation}

\noindent where \(F\&R(\cdot)\) denotes the feature fusion and reconstruction module, and \(I_{F}\) represents the fused image.

\begin{figure}[!t]
\centering
\includegraphics[width= \columnwidth]{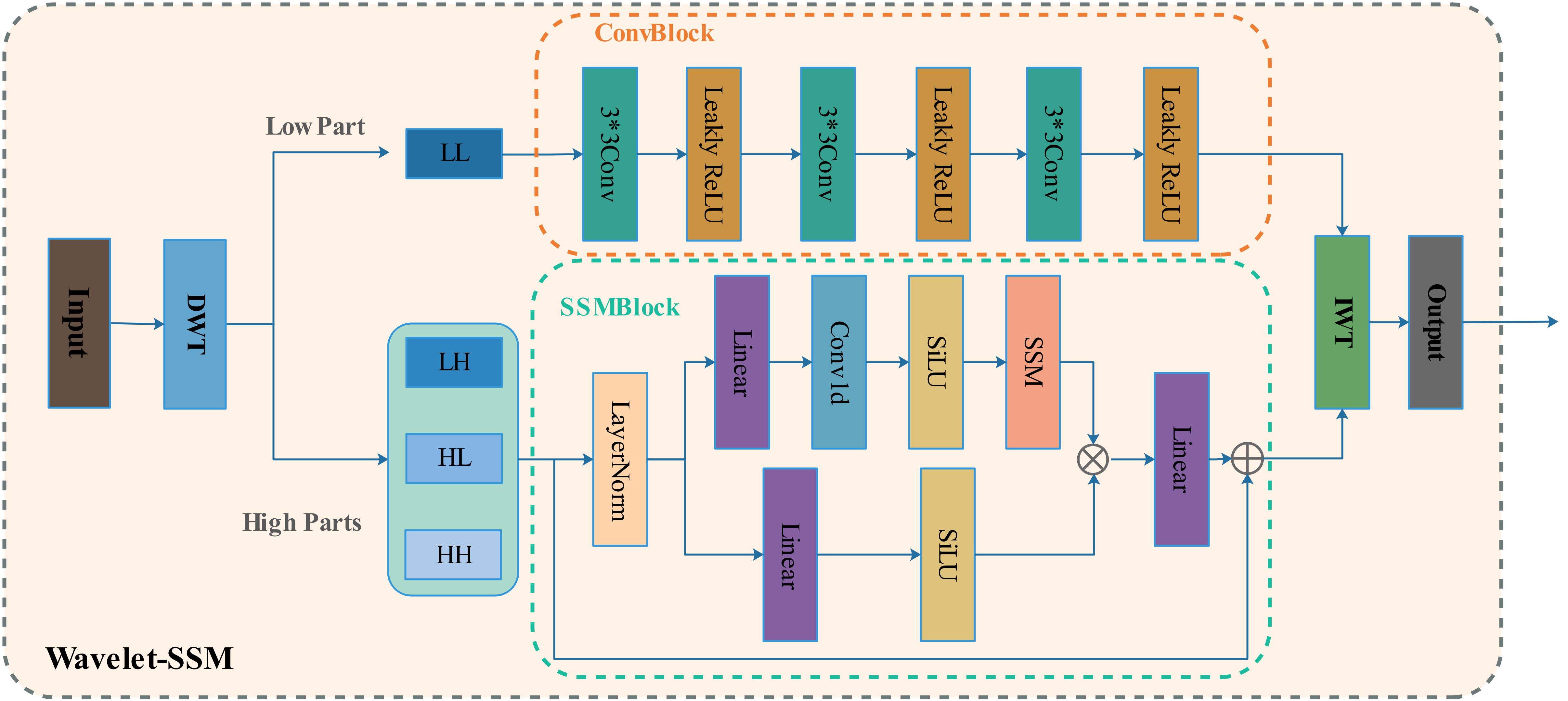}
\caption{The detailed structure of the Wavelet-Mamba in Fig.\ref{Architecture}(a).}
\label{SSM-Wavelet}
\end{figure}

\subsection{Wavalet-SSM Feature Extraction}
\subsubsection{\textbf{State Space Model}}
The State Space Model (SSM) \cite{xie2024fusionmamba} maps the input sequence \(x(t)\) to a latent state representation \(h'(t)\) using state equations and generates the predicted output \(y(t)\) through output equations:

\begin{equation}
\label{eq:ssm}
h'(t) = Ah(t)+Bx(t),
y(t) = Ch(t)+Dx(t)  
\end{equation}

\noindent where \(h'(t)\) represents the current state, \(h(t)\) denotes the previous state, and \(A\), \(B\), \(C\), and \(D\) are the model parameters. To integrate the SSM into deep learning, the Zero-Order Hold (ZOH) technique is utilized for discretization. Vision Mamba replaces the traditional self-attention mechanism with a bidirectional SSM, leveraging it to model sequential data and efficiently capture long-range dependencies.

\subsubsection{\textbf{Architecture}}
The structure of WFE is shown in Fig.\ref{Architecture}(a), and the process is described as follows:

\begin{equation}
\label{eq:FIM2}
\begin{aligned}
f' &= f+WM(LN(f)) \\
f_{out} &= f'+GAM(LN(f'))
\end{aligned}
\end{equation}

\noindent where \(f\) represents the input to the WFE, \(LN(\cdot)\) denotes the layer normalization operation, \(WM(\cdot)\) represents the Wavelet-Mamba operation, \(GAM(\cdot)\) denotes the gated attention mechanism, and \(f_{out}\) is the output of the WFE.

\textbf{Wavelet-Mamba:} The detailed structure of the Wavelet-Mamba is illustrated in Fig.\ref{SSM-Wavelet}. A two-dimensional discrete wavelet transform (2D-DWT) is applied to the input features, decomposing them into low-frequency and high-frequency components:
\begin{equation}
\label{eq:2ddwt}
LL,LH,HL,HH = 2DDWT(f_{WM}^{in})
\end{equation}

\noindent where, \( f_{WM}^{in} \) represents the input to the Wavelet-Mamba block. The \(LL\) component corresponds to low-frequency features, while \(LH\), \(HL\), and \(HH\) represent high-frequency components. Low-frequency features are processed using convolutional layers, whereas high-frequency features are refined via SSM to capture long-range dependencies. This approach facilitates the comprehensive extraction of global features and local details in the frequency domain. Finally, an inverse transformation reconstructs these features back into the spatial domain.

\begin{equation}
\label{eq:FIM}
\begin{aligned}
f_{lf} &= ConvBlock(LL) \\
f_{hf} &= SSMBlock(\{LH,HL,HH\})\\
f_{WM}^{out} &= 2DIWT(concat(f_{lf},f_{hf}))
\end{aligned}
\end{equation}

\noindent where \(f_{lf}\) and \(f_{hf}\) represent the feature maps produced by the convolution operations \(ConvBlock(\cdot)\) and the SSM process \(SSMBlock(\cdot)\), respectively. The final output vector \(f_{WM}^{out}\) is obtained by concatenating (\(concat(\cdot)\)) these feature maps, followed by the inverse wavelet transformation \(2DIWT(\cdot)\).

\begin{figure}[!t]
\centering
\includegraphics[width= \columnwidth]{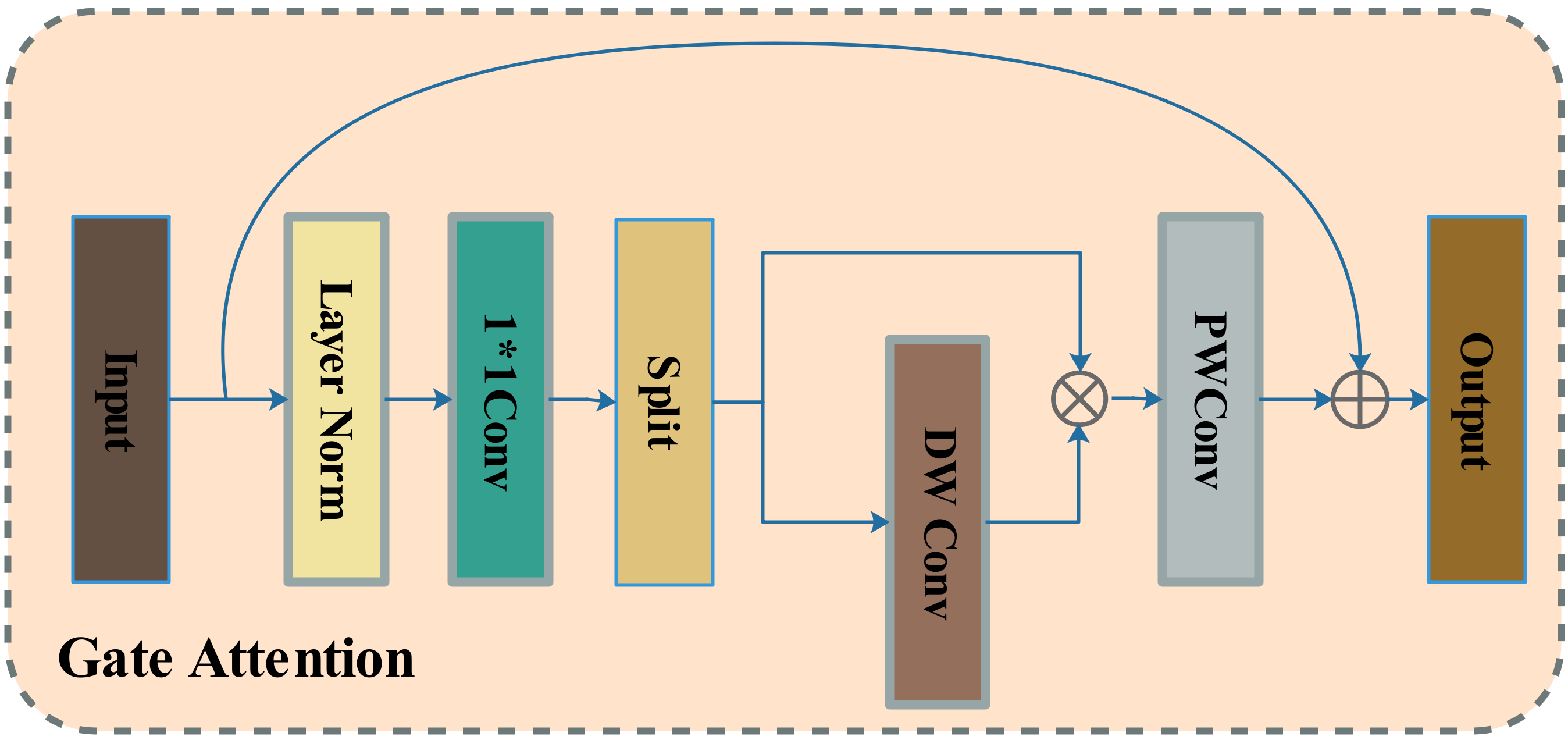}
\caption{The detailed structure of the Gated Attenition in Fig.\ref{Architecture}(a).}
\label{gateattention}
\end{figure}

\textbf{Gated Attention Mechanism:} Inspired by \cite{Wang2024gate}, we propose a gated mechanism utilizing depthwise separable convolutions (Fig.\ref{gateattention}) for dynamic information control. Depthwise convolutions capture weights based on local relationships, facilitating efficient feature extraction while minimizing redundancy and reducing parameter overhead.
\begin{equation}
\label{eq:dgam}
\begin{aligned}
&f_{att},w_{att}=Split(Conv_{1}^{C,2C}(LN(f_{GAM}^{in})))\\
&f_{GAM}^{out} = f_{GAM}^{in}+PWConv(f_{att} \otimes DWConv(w_{att}))
\end{aligned}
\end{equation}

\noindent where, \( f_{GAM}^{in} \) represents the input, \( Conv_{1}^{C,2C}(\cdot) \) denotes a 1\(\times\)1 convolution that maps \(C\) channels to \(2C\) channels, and \( Split(\cdot) \) performs channel-wise splitting to produce the feature \( f_{att} \) and the weight \( w_{att} \). The symbol \( \otimes \) represents element-wise multiplication, while \( DWConv(\cdot) \) and \( PWConv(\cdot) \) denote depth-wise and point-wise convolutions, respectively. The output of the GAM module is \( f_{GAM}^{out} \).

\subsection{Cross-modal Attention Feature Modulation}
The Cross-modal Attention Feature Modulation (CAFM) is shown in Fig.\ref{Architecture}(b). It enhances single-modal features by leveraging cross-modal complementary information, thereby improving feature expressiveness while preserving the original information. For the inputs \( f_{vi}^{in}, f_{ir}^{in} \in R^{C' \times H' \times W'} \), the padding process is defined as:

\begin{equation}
\label{eq:tokensplit}
\begin{aligned}
    f_{vi}^{pf} = [f_{vi}^{1},f_{ir}^{1},f_{vi}^{2},f_{ir}^{2},...,f_{vi}^{n},f_{ir}^{n}]\\
    f_{ir}^{pf} = [f_{ir}^{1},f_{vi}^{1},f_{ir}^{2},f_{vi}^{2},...,f_{ir}^{n},f_{vi}^{n}]
\end{aligned}
\end{equation}

\noindent where \( f_{vi}^{i}, f_{ir}^{i} \in R^{1 \times H' \times W'} \) are the \(i\)-th channels of \( f_{vi}^{in} \) and \( f_{ir}^{in} \), while \( f_{vi}^{pf}, f_{ir}^{pf} \in R^{2C' \times H' \times W'} \) are the padded features. Meanwhile, a learnable weight matrix and linear projection map the padded features for modality enhancement and adjustment of dimensions.

\begin{equation}
\label{eq:token}
\begin{aligned}
    f_{vi}^{out} = Proj_{2C'}^{C'}(\mathcal{I}^{vi} \otimes f_{vi}^{pf})\\
    f_{ir}^{out} = Proj_{2C'}^{C'}(\mathcal{I}^{ir} \otimes f_{ir}^{pf})
\end{aligned}
\end{equation}

\noindent where \( \mathcal{I}^{vi}, \mathcal{I}^{ir} \in R^{2C' \times 1 \times 1} \) represent the weights parameters, and \( Proj_{2C'}^{C'}(\cdot) \) denotes a linear projection reducing \( 2C' \) channels to \( C' \). This process yields \( f_{vi}^{out} \) and \( f_{ir}^{out} \), embedding original modality and cross-domain complement.

\begin{figure*}[!t]
\centering
\includegraphics[width=1.0 \textwidth]{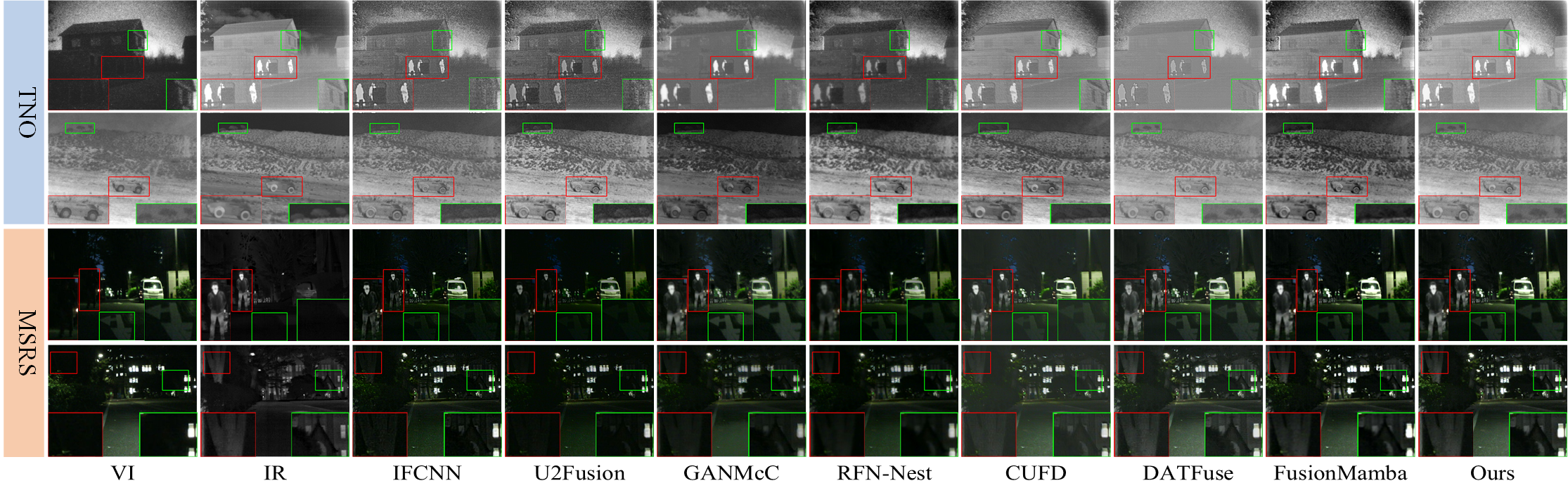}
\caption{Qualitative comparison of W-Mamba and seven comparative methods on TNO (top two rows) and MSRS  (bottem two rows) dataset. We highlight two key regions in the image with \textcolor{red}{red} and \textcolor{green}{green} bounding boxes, respectively, and magnify them to simplify the visual comparison process.}
\label{Qualitative}
\end{figure*}

\subsection{Loss function}
We train W-Mamba by intensity and gradient loss to preserve structural textures and regulate intensity. The overall loss function is defined as:

\begin{equation}
\label{eq:lossall}
  \mathcal{L}_{all} = \lambda_{1}\mathcal{L}_{int} + \lambda_{2}\mathcal{L}_{grad}
\end{equation}
\noindent where $\lambda_{1}$ and $\lambda_{2}$ are weighting parameters used to balance the two sub-loss functions. The intensity loss encourages the training network to preserve more meaningful pixel intensity information. $\mathcal{L}_{int}$ is defined as shown in Eq.~\ref{eq:lossint}.

\begin{equation}
  \mathcal{L}_{int} = \frac{1}{HW}||I_{F}-max(I^{ir}, I^{vi})||_{1}
  \label{eq:lossint}
\end{equation}

\noindent in Eq.~\ref{eq:lossint}, $max(\cdot)$ denotes the element-wise maximum operation. Additionally, The gradient loss is designed to retain as much texture detail as possible from both modalities. $\mathcal{L}_{grad}$ is defined in Eq.~\ref{eq:lossgrad}.

\begin{equation}
  \mathcal{L}_{grad} = \frac{1}{HW}|||\nabla I_{F}|-max(|\nabla I^{ir}|, |\nabla I^{vi}|)||_{1}
  \label{eq:lossgrad}
\end{equation}

\noindent where $\nabla$ is Sobel gradient operator, $max(\cdot)$ denotes the element-wise maximum selection.

\begin{figure*}
\centering
\includegraphics[width=1.0 \textwidth]{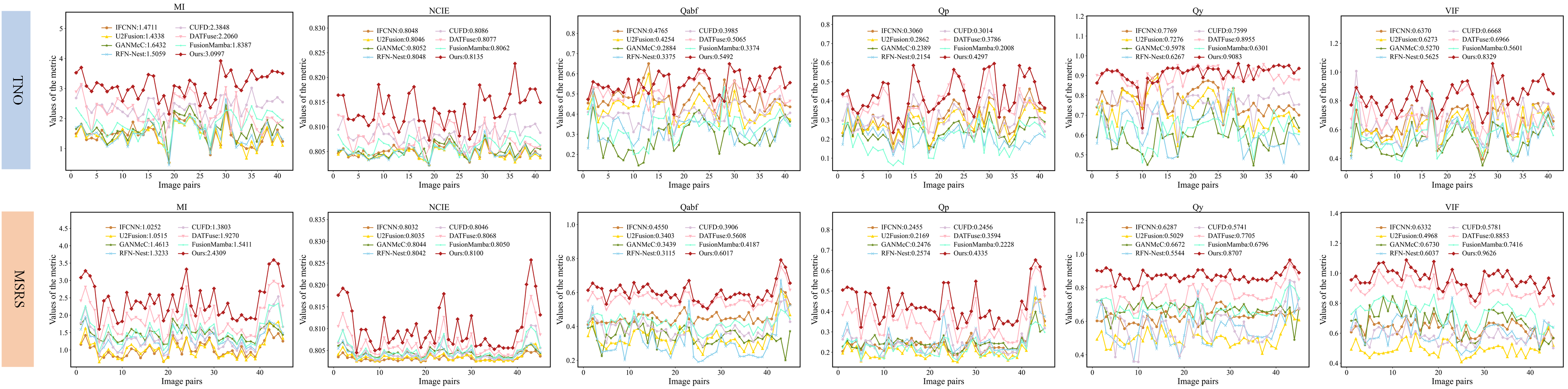}
\caption{Quantitative comparison of W-Mamba and seven comparison methods on TNO\cite{toet2017tno} and MSRS\cite{tang2022piafusion} dataset. Our method is represented by the red line. Mean values of each method are shown in each legend. For the 6 metrics used for comparison, a higher value indicates superior performance.}
\label{fig:Quantitative}
\end{figure*}

\section{Experiment Results and Analysis}
\subsection{Experiment Configurations}
\textbf{Implementation details}: We randomly cropped paris of image patches with dimensions 128 \(\times\) 128 from the TNO \cite{toet2017tno}, MSRS \cite{tang2022piafusion} and LLVIP \cite{jia2021llvip} datasets, and normalizing their pixel intensity to the range $[0, 1]$. The Adam optimizer was employed with a learning rate of \(2.5\times 10^{-5}\). The batch size was set to 24. The hyperparameters of the loss function, \(\lambda_{1}\) and \(\lambda_{2}\) were empirically set to 10 and 1, respectively. All experiments were conducted on a NVIDIA GeForce RTX 4090D GPU. Finally, we evaluate W-Mamba on the TNO \cite{toet2017tno} and MSRS \cite{tang2022piafusion} datasets to assess the performance.

\textbf{Evaluation metrics}: We selected AE-based methods RFN-Nest \cite{li2021rfn} and CUFD \cite{xu2022cufd}, CNN-based methods U2Fusion \cite{xu2020u2fusion} and IFCNN \cite{zhang2020ifcnn}, GAN-based method GANMcC \cite{ma2020ganmcc}, Transformer-based method DATFuse \cite{tang2023datfuse}, and Mamba-based method FusionMamba \cite{xie2024fusionmamba}. Additionally, we employed six widely recognized metrics for quantitative evaluation, including two information-theory-based metrics: mutual information (MI) \cite{qu2002information} and nonlinear correlation information entropy (NCIE) \cite{wang2005nonlinear}; two image feature-based metrics: gradient-based similarity measurement (Q\(_{abf}\)) \cite{xydeas2000objective} and phase congruency (Q\(_{p}\)) \cite{zhao2007performance}; one structural similarity-based metric: Yang's metric (Q\(_{y}\)) \cite{li2008novel}; and one human visual perception-based metric: visual information fidelity (VIF) \cite{han2013new}.

\begin{table}[!t]
\footnotesize
\centering
\caption{Quantitative comparisons between W-Mamba and seven methods on the TNO\cite{toet2017tno} and MSRS\cite{tang2022piafusion} datasets. The top three performers for each metric are highlighted in \textcolor{red}{red}, \textcolor{blue}{blue}, and \textcolor{green}{green}, respectively. \(\uparrow\) indicates that higher values correspond to better performance.}
\resizebox{  \columnwidth}{!}{ 
    \begin{tabular}{ccccccc}
    \toprule
        \textbf{TNO} & \textbf{MI$\uparrow$} & \textbf{NCIE$\uparrow$} & \textbf{Q$_{abf}\uparrow$} & \textbf{Q$_{p}\uparrow$} & \textbf{Q$_{y}\uparrow$} & VIF$\uparrow$\\
    \midrule
    IFCNN\cite{zhang2020ifcnn} & 1.4711 & 0.8048 &\textcolor{green}{0.4765} & \textcolor{green}{0.3060} & \textcolor{green}{0.7769} &0.6370 \\
    U2Fusion\cite{xu2020u2fusion} & 1.4338 & 0.8046 & 0.4254 & 0.2862 & 0.7276 & 0.6273\\
    GANMcC\cite{ma2020ganmcc} & 1.6432 & 0.8052 & 0.2884 & 0.2389 & 0.5978 &0.5270 \\
    RFN-Nest\cite{li2021rfn} & 1.5059 & 0.8048 & 0.3375 & 0.2154 & 0.6267 & 0.5625\\
    CUFD\cite{xu2022cufd} & \textcolor{blue}{2.3848} & \textcolor{blue}{0.8086} & 0.3985 & 0.3014 & 0.7599 & \textcolor{green}{0.6668}\\
    DATFuse\cite{tang2023datfuse} & \textcolor{green}{2.2060} & \textcolor{green}{0.8077} & \textcolor{blue}{0.5065} & \textcolor{blue}{0.3786} & \textcolor{blue}{0.8955} & \textcolor{blue}{0.6966}\\
    FusionMamba\cite{xie2024fusionmamba} & 1.8387 & 0.8062 & 0.3374 & 0.2008& 0.6301 &0.5601 \\
    \rowcolor[rgb]{0.9,0.9,0.9}$\star$\textbf{Ours} & \textcolor{red}{3.0997} & \textcolor{red}{0.8135} & \textcolor{red}{0.5492} & \textcolor{red}{0.4297} & \textcolor{red}{0.9083} &\textcolor{red}{0.8329} \\
    \midrule
    \textbf{MSRS} & \textbf{MI$\uparrow$} & \textbf{NCIE$\uparrow$} & \textbf{Q$_{abf}\uparrow$} & \textbf{Q$_{p}\uparrow$} & \textbf{Q$_{y}\uparrow$} & VIF$\uparrow$\\
    \midrule
    IFCNN\cite{zhang2020ifcnn} & 1.0252 & 0.8032 & \textcolor{green}{0.4550} & 0.2455 & 0.6287 &0.6332 \\
    U2Fusion\cite{xu2020u2fusion} & 1.0515 & 0.8035 & 0.3403 & 0.2169 & 0.5029 & 0.4968\\
    GANMcC\cite{ma2020ganmcc} & 1.4613 & 0.8044 & 0.3439 & 0.2476 & 0.6672 &0.6730 \\
    RFN-Nest\cite{li2021rfn} & 1.3233 & 0.8042 & 0.3115 & \textcolor{green}{0.2574} & 0.5544 & 0.6037\\
    CUFD\cite{xu2022cufd} & 1.3803 & 0.8046 & 0.3906 & 0.2456 & 0.5741 & 0.5781\\
    DATFuse\cite{tang2023datfuse} & \textcolor{blue}{1.9270} & \textcolor{blue}{0.8068} & \textcolor{blue}{0.5608} & \textcolor{blue}{0.3594} & \textcolor{blue}{0.7705} & \textcolor{blue}{0.8853}\\
    FusionMamba\cite{xie2024fusionmamba} & \textcolor{green}{1.5411} & \textcolor{green}{0.8050} & 0.4187 & 0.2228& \textcolor{green}{0.6796} &\textcolor{green}{0.7417} \\
    \rowcolor[rgb]{0.9,0.9,0.9}$\star$\textbf{Ours} & \textcolor{red}{2.4309} & \textcolor{red}{0.8100} & \textcolor{red}{0.6017} & \textcolor{red}{0.4335} & \textcolor{red}{0.8707} &\textcolor{red}{0.9626} \\
    \bottomrule
    \end{tabular}
}
    \label{tab:quantitative}
\end{table}

\begin{figure*}[!t]
\centering
\includegraphics[width=\textwidth]{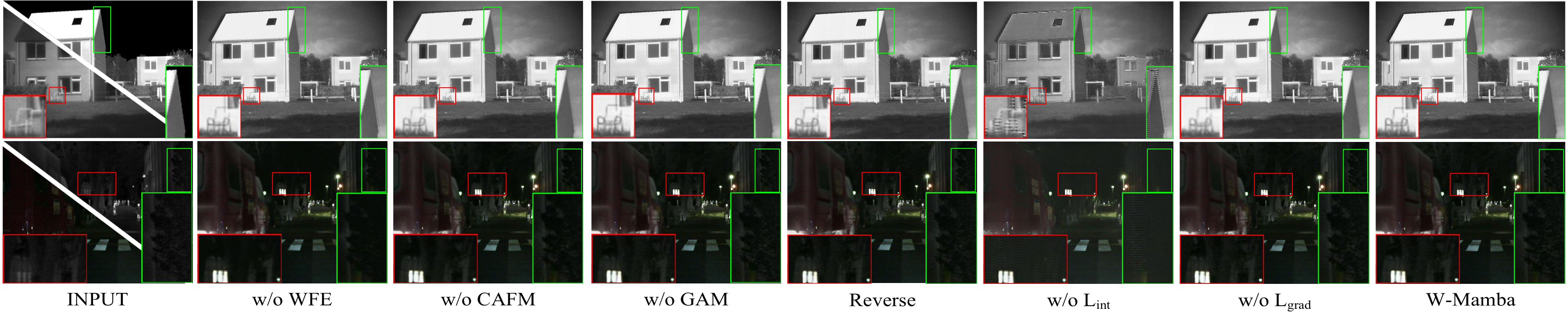}
\caption{Qualitative comparison of W-Mamba ablation experiment. We highlight two key regions in the image with \textcolor{red}{red} and \textcolor{green}{green} bounding boxes, respectively, and magnify them to simplify the visual comparison process.}
\label{fig:Qualitative_Ablation}
\end{figure*}

\subsection{Experiment results}
\textbf{Quantitative Analysis}: The quantitative results are presented in Tab.\ref{tab:quantitative} and Fig.\ref{fig:Quantitative}. Our approach consistently outperforms the others on six evaluation metrics. The results demonstrate that W-Mamba effectively extracts complementary information from source images, producing fused images with enriched information, comprehensive features, and strong structural similarity.

\textbf{Qualitative Analysis}: The fusion results of W-Mamba and other methods on the TNO \cite{toet2017tno} and MSRS \cite{tang2022piafusion} are shown in Fig.\ref{Qualitative}. IFCNN, U2Fusion, and FusionMamba in Fig.\ref{Qualitative} introduced considerable noise, such as the green box in the first row, which caused the loss of edge information. The green box in the second row indicates that our method has better fusion performance in two source images where the foreground and background brightness are opposite. In the infrared images with grayscale or RGB images fusion, DATFuse has low contrast in the grayscale fusion results, such as the window detail to the left of the green box in the first row of Fig.\ref{Qualitative}. However, CUFD has low contrast in RGB image fusion, resulting in unclear visible and infrared edge information. Overall, W-Mamba can utilize complementary information from two modalities, can fusion results with clear edges, prominent targets, thus alignment well with visual perception.

\begin{table}[!t]
\footnotesize
\centering
\caption{Quantitative comparison of ablation experiments. \textcolor{red}{Red} represents the best result. \textbf{Reverse} represents applying SSM to low-frequency part. $\uparrow$ indicates that higher metrics the better performance.}
\label{tab:ablation}
\resizebox{ \columnwidth}{!}{
\begin{tabular}{cccccccc}
\toprule
 & \textbf{MI$\uparrow$} & \textbf{NCIE$\uparrow$} & \textbf{Q$_{abf}$$\uparrow$} & \textbf{Q$_{p}$$\uparrow$} & \textbf{Q$_{y}$$\uparrow$} & \textbf{VIF$\uparrow$} \\
\midrule
 w/o WFE & 2.5857 & 0.8101 & 0.4685 &  0.3337 & 0.8129 & 0.6946 \\
 w/o CAFM & 2.8825 & 0.8119 & 0.5096 & 0.3812 & 0.8795 & 0.7723 \\
 w/o GAM & 3.0019 & 0.8128 & 0.5363 & 0.4158 & 0.9013 & 0.8138 \\
 Reverse & 2.7409 & 0.8110 & 0.4932 & 0.3645 & 0.8619 & 0.7475 \\
\midrule
 w/o $L_{int}$ & 1.3338 & 0.8047 & 0.4649 & 0.3171 & 0.4276 & 0.3478 \\
 w/o $L_{grad}$ & 2.6198 & 0.8102 & 0.4497 & 0.3281 & 0.8198 &0.6838 \\
 
\midrule
 \rowcolor[rgb]{0.9,0.9,0.9}$\star$\textbf{W-Mamba} & \textcolor{red}{3.0997} & \textcolor{red}{0.8135} & \textcolor{red}{0.5492} & \textcolor{red}{0.4297} & \textcolor{red}{0.9083} & \textcolor{red}{0.8329} \\
\bottomrule
\end{tabular}
}
\end{table}

\subsection{Ablation Study}
\textbf{Quantitative Analysis}: The ablation studies are conducted  from two perspectives: \textbf{network architecture} ablation (w/o WFE, CAFM, GAM and reverse SSM frequency band) and \textbf{loss function} ablation (w/o $L_{int}$ and $L_{grad}$). As shown in Tab.\ref{tab:ablation}. The remove of WFE, CAFM and GAM resulted in a decline in all six evaluation indicators, demonstrating the effectiveness of frequency domain information and global information, and demonstrating the importance of complementary information between modalities. Furthermore, when SSM is applied to low-frequency signals in wavelet transform, each indicator experiences a degradation, indicating that SSM can better extract complementary features in high-frequency parts. Finally, the removal of the intensity loss function and gradient loss function, $Q_{abf}$ and $Q_{p}$ will significantly decrease, indicating that the gradient loss function is crucial in preserving details.

\textbf{Qualitative Analysis}: The qualitative results are presented in Fig.\ref{fig:Qualitative_Ablation}, revealing the following observations: 1) After removing WFE, the feature extraction performance declines and resulting in significant performance degradation, which is artifacts in the green box of the first row. The intensity information is lost in the red box of the second row, resulting in blurred edges after fusion. 2) Removing CAFM presents challenges in enhancing complementary features, resulting in edge artifacts in the green box of the first row. 3) The removal of GAM partially reduces the ability to capture local relationships in the source images, leading to blurred leaf details in the green box of the second row. 4) When SSM is applied to the low-frequency components, the green box in the first row displays pseudo edges at the edge of the roof, indicating that SSM is effective in capturing long-range dependencies in the high-frequency. 5) After removing the intensity loss function, the fusion result fails to produce appropriate intensity, and some distortion occurs such as the red and green boxes in the first row. 6)After removing the gradient loss function, the edges of the fused image generated at the edge of the chair in the first row are blurred.

\begin{table}[!t]
\caption{Quantitative comparison of object detection on the MSRS\cite{tang2022piafusion} dataset. The best-performing method is highlighted in \textcolor{red}{red}. \(\uparrow\) indicates that higher values correspond to better performance.}
\footnotesize
\label{tab:detection}
\footnotesize
\centering
\resizebox{ \columnwidth}{!}{ 
    \begin{tabular}{ccccc}
    \toprule
        Method & \textbf{mAP@0.65} $\uparrow$& \textbf{mAP@0.85} $\uparrow$& \textbf{mAP@[0.5,0.95] $\uparrow$}\\
    \midrule
    IFCNN\cite{zhang2020ifcnn} & 0.806 & 0.473 & 0.621 \\
    U2Fusion\cite{xu2020u2fusion} & 0.837 & 0.498 & 0.636 \\
    GANMcC\cite{ma2020ganmcc} & 0.842 & 0.534 & 0.656\\
    RFN-Nest\cite{li2021rfn} & 0.775 & 0.476 & 0.593\\
    CUFD\cite{xu2022cufd} & \textcolor{red}{0.849} & 0.507 & 0.649 \\
    DATFuse\cite{tang2023datfuse} & 0.843 & 0.497 & 0.645\\
    FusionMamba\cite{xie2024fusionmamba} & 0.797 & 0.497 & 0.622\\
    \rowcolor[rgb]{0.9,0.9,0.9}$\star$\textbf{Ours} & 0.848 & \textcolor{red}{0.565} & \textcolor{red}{0.659}\\
    \bottomrule
    \end{tabular}
}
\end{table}

\subsection{Downstream application}
To demonstrate the superiority of W-Mamba in downstream tasks, we fused 80 pairs of labeled images from the MSRS dataset with YOLOv5. Tab.\ref{tab:detection} presents the mean average precision (mAP) under various intersection over union (IoU) thresholds. The results indicate that our method achieves the highest performance in mAP@[0.5, 0.95], demonstrating its robust mAP performance. This further validates the effectiveness of W-Mamba.

\section{Discussion and conclusion}
In this paper, We propose Wavelet-Mamba (W-Mamba). The key innovation is Wavelet-SSM Feature Extraction module, which combines frequency domain information with SSM. We also propose a gated attention mechanism to extract details near the edges. Additionally, we propose cross-modal attention feature modulation, which dynamically fuses cross-modal features. The experimental results show that our method is superior to existing IVIF methods and enhances visual perception in the fused images.

\bibliographystyle{IEEEbib}
\bibliography{icme2025references}

\end{document}